\pdfoutput=1

\documentclass[11pt]{article}

\usepackage{authblk}
\usepackage{emnlp2021}

\usepackage{times}
\usepackage{latexsym}
\usepackage{float}
\newcommand{\etc}{\emph{etc.\ }}
\newcommand{\eg}{\emph{e.g.},\ }
\usepackage{graphicx}
\usepackage{natbib}

\restylefloat{table}
\usepackage{placeins}
\usepackage[T1]{fontenc}

\usepackage[utf8]{inputenc}

\usepackage{microtype}

\title{{C}o{P}{H}{E}: {A} {C}ount-{P}reserving {H}ierarchical {E}valuation {M}etric in {L}arge-{S}cale {M}ulti-{L}abel {T}ext {C}lassification}

\author[1]{Mat\'{u}\v{s} Falis} 
\author[2]{Hang Dong} 
\author[1]{Alexandra Birch} 
\author[1,3,4]{Beatrice Alex}
\affil[1]{Institute for Language, Cognition and Computation, University of Edinburgh}
\affil[2]{Centre for Medical Informatics, Usher Institute, University of Edinburgh}
\affil[3]{School of Literatures, Languages and Cultures, University of Edinburgh}
\affil[4]{Edinburgh Futures Institute, University of Edinburgh}
\affil[ ]{\texttt {\{s1206296, Hang.Dong, a.birch, balex\}@ed.ac.uk}}
\begin{document}

\maketitle
\begin{abstract}
  \noindent{Large-Scale Multi-Label Text Classification (LMTC) includes tasks with hierarchical label spaces, such as automatic assignment of ICD-9 codes to discharge summaries. Performance of models in prior art is evaluated with standard precision, recall, and $F_1$ measures without regard for the rich hierarchical structure. In this work we argue for hierarchical evaluation of the predictions of neural LMTC models. With the example of the ICD-9 ontology we describe a structural issue in the representation of the structured label space in prior art, and propose an alternative representation based on the depth of the ontology. We propose a set of metrics for hierarchical evaluation using the depth-based representation. We compare the evaluation scores from the proposed metrics with previously used metrics on prior art LMTC models for ICD-9 coding in MIMIC-III. We also propose further avenues of research involving the proposed ontological representation.}
\end{abstract}

\section{Introduction}

The Large-Scale Multi-Label Text Classification (LMTC) is a multi-label document classification task with a large (in the order of thousands) and potentially structured label space. LMTC tasks with structured label spaces can be found in the domain of medicine, legislation, product categorisation, \etc Individual label spaces may follow different labelling rules, \eg the \emph{EURLEX} \cite{chalkidis2019large} dataset's labels can be any node from the EUROVOC tree of concepts, while \emph{AMAZON13K} \cite{mcauley2013hidden} assumes a dense assignment, where if a predicted (leaf) node's truth value is positive, so is the truth value of all its ancestors -- also known as the \emph{true path rule}. A representative dataset for medical LMTC are \emph{MIMIC-III} \cite{johnson2016mimic} discharge summaries. These are weakly labelled with leaf nodes of the ICD-9 ontology\footnote{\url{https://www.cdc.gov/nchs/icd/icd9cm.htm}}. ICD-9 is a tree-structured ontology of medical conditions and procedures.

Since the release of the MIMIC-III dataset there have been several attempts at training neural models for automated coding of medical documents \cite {Mullenbach_Wiegreffe_Duke_Sun_Eisenstein_2018, Rios_Kavuluru_2018,Falis_2019, Chalkidis_Fergadiotis_Kotitsas_Malakasiotis_Aletras_Androutsopoulos_2020, Dong_Suarez-Paniagua_Whiteley_Wu_2020}. While some prior art has made use of ontological structure \cite{Rios_Kavuluru_2018, Falis_2019, Manginas_Chalkidis_Malakasiotis_2020}, the task has mostly been treated as a flat prediction of the assigned leaves. This is reflected in the evaluation metrics that are used across previous work -- precision, recall and $F_1$ score on flat predictions. These metrics applied to flat vectors disregard the rich ontological structure. A notable exception is \citet{Manginas_Chalkidis_Malakasiotis_2020}  -- they fine-tune a BERT \cite{devlin2018bert} model, such that different layers within the model learn to represent the different depths of the target label space. Each layer undergoes flat evaluation with respect to labels on its depth.

The aim of this study is to review the use of ontological structure within the prior LMTC art with a focus on automated ICD-9 coding, point out issues in hierarchy representation in prior approaches, describe methods more suited for addressing the structured label space including a hierarchical evaluation metric, and propose further avenues of research utilising structured label spaces. Our implementation of evaluation and the representation of ICD-9's graph are available to the community\footnote{\url{https://github.com/modr00cka/CoPHE}}.

\section{Background}

Within a structured label space some labels are inherently closer to one another than to others - \eg in ICD-9 \textit{425.0 Endomyocardial fibrosis} is closer to \textit{425.3 Endocardial fibroelastosis} than to  \textit{305.1 Tobacco use disorder}. Flat prediction and standard precision/recall/$F_1$ score (from hereon referred to as \emph{standard metrics}) of individual prediction level (leaf) codes treat all mispredictions equally -- \eg having \emph{425.0} be mispredicted as \emph{425.3} is penalised the same way as mispredicting it as \emph{305.1}. This phenomenon has been addressed in information extraction (IE) by \citet{Maynard_Peters_Li} through the use of distance metrics. The IE setting assumes both the gold standard and predictions to be associated with specified spans within the input text. This means an individual prediction can be associated with a true label, allowing direct comparison between them. 

The LMTC setting uses weak labels -- predictions and true labels appear on the document level, without exact association to spans within the text. Due to the absence of information regarding associated spans, direct links between individual predictions and true labels do not exist. Label comparison is performed on full vectors representing multiple labels, hence the IE approach is not directly usable. \citet{Kosmopoulos_Partalas_Gaussier_Paliouras_Androutsopoulos_2015} address hierarchical label spaces in document classification with \emph{set-based} measures. The gold standard and prediction vectors are extended to include ancestor nodes within the hierarchical label-space and augmented according to its structure, and the true-path rule.

Let $X = \{x_i|1, ..., M\}$ and let $Y = \{y_i|1, ..., N\}$ represent the set of predicted and true codes for a certain document respectively. Assume we have access to an augmentation function $An_j(x)$ which returns the ancestors of $x$ up to the $j^{th}$ order. 

Let $X_{aug} = X \cup \{An_j(x_i)|1, ..., M\}$ and let $Y_{aug} = Y \cup \{An_j(y_i)|1, ..., N\}$ represent the set of predicted and gold ancestor codes for $X$ and $Y$ respectively. Standard metrics can then be applied to the $X_{aug}$ and $Y_{aug}$ sets.  A correct assignment of predicted lower-level (leaf) codes results in correct assignment of their ancestors. In the case of incorrect prediction-level  assignments, the closer a mismatched predicted code is to the gold standard leaf within the ontology, the more matches will occur across the levels of the hierarchy.

 On the leaf level each code appears at most once per document. Duplicates can occur when $An_j(x)$ produces the same ancestor for multiple codes. As $X_{aug}$ is a set, duplicates are removed. Hence, the set-based approach captures whether an ancestor is present, but not how many of its descendants were predicted. This results in loss of information regarding over-/under- predictions of classes on the ancestral level. Over- and under-prediction is a valuable phenomenon to track, particularly if the label-space includes inexplicit rules -- for instance, for some nodes only a single descendant can be predicted at a time, as individual siblings are mutually exclusive (\eg a patient can be assigned at most one of codes \emph{401.0}, \emph{401.1}, and \emph{401.9}, which represent malignant, benign, and unspecified hypertension respectively -- concepts that are mutually exclusive). Furthermore, retaining this numeric data on ancestral levels enables analyses on higher levels, \eg performance of a family of codes in the case of a semi-automated code-assignment application. For this reason we propose using a metric that retains the descendant counts for these ancestor codes.

To correctly define the augmentation function we need to ensure our representation of the hierarchy fits the setting. Previous work involving the ICD-9 hierarchy \cite{Rios_Kavuluru_2018, Falis_2019, Chalkidis_Fergadiotis_Kotitsas_Malakasiotis_Aletras_Androutsopoulos_2020} represents it through the relation of direct ancestry considering parents and grandparents of the leaf nodes. As the ICD-9 has leaves at different depths, this representation results in structural issues, such as one code being both in the position of a parent and a grandparent for different leaves. For instance, code \textit{364} has a parent relation to the leaf \textit{364.3} and grandparent to leaf \textit{364.11} (Figure \ref{fig:sub2}). This poses an issue to aggregation and evaluation. We aim to address this issue by producing a representation of the hierarchy through the levels of the label space with each level representing all the nodes at a certain depth in the tree structure.

\section{Method}

\subsection{Augmentation}
To implement level-based augmentation, we first define the first three layers of the ICD-9 hierarchy on which leaves appear. An ICD-9 code (\eg 364.11) consists of a ``category'' (part of the code appearing prior to the decimal point, \eg 364) and ``etiology'' (appearing after the decimal point, \eg 11). The etiology can be represented with up to two digits. We define the basic levels of the hierarchy (encapsulating all the labels in MIMIC-III) as follows: codes with double digit etiology ($e_2$); codes of single digit etiology ($e_1$); codes described only with ``category'' (no etiology, $e_0$). Augmentation can be performed up to a higher user-defined level within the hierarchy by adding further layers representing chapters within the ontology.

The originally flat predicted and true labels are divided into their respective layers in the hierarchy. If a code appears in a level lower than the maximum level set by the user, the truth value of the code is propagated to its direct ancestor through augmentation. The propagation can be interpreted either as a truth value (\emph{binary}) or the number of descendant leaves present (\emph{count-preserving}). The binary interpretation of the propagation results in the ancestor holding the truth value of the logical OR operation on its children. This mimics the set-based approaches described by  \citet{Kosmopoulos_Partalas_Gaussier_Paliouras_Androutsopoulos_2015}. The count-preserving interpretation sets the value of the ancestor to be the sum of the values of its descendants. Through retaining the numeric information, the count-preserving interpretation allows us to track over- or under-prediction within a family of codes.

\subsection{Hierarchical Evaluation}
\begin{figure}
  \centering
  \includegraphics[width=\linewidth]{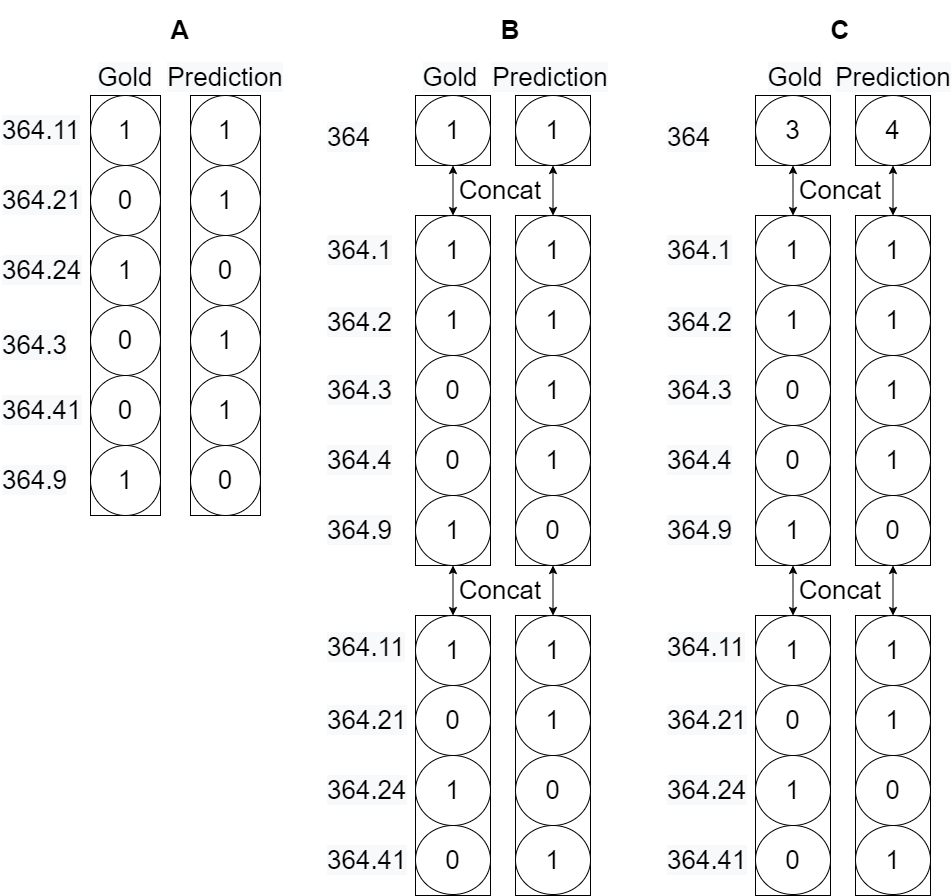}
  \caption{A comparison between three styles of evaluation: \textbf{(A)} Evaluation performed only on leaf nodes (no use of hierarchical relations); \textbf{(B)} Set-Based Hierarchical Evaluation  \cite{Kosmopoulos_Partalas_Gaussier_Paliouras_Androutsopoulos_2015}. Four descendants of the node \emph{364} are predicted, and three appear in the gold standard. This is reflected in the middle-level vector, but the information is lost in the top level; \textbf{(C)} Count-Preserving Hierarchical Evaluation. The numeric information of predictions and gold labels are preserved on higher levels.}
  \label{fig:comparison}
 \end{figure}

To produce hierarchical precision, recall, and $F_1$ up to a certain level within the hierarchy, the vectors from each layer are combined for the predictions and true labels respectively. In the case of set-based interpretation (Figure \ref{fig:comparison} B), standard metrics can be directly applied to the augmented vectors.

Since the notions of true positive ($TP$), false negative ($FN$), and false positive ($FP$) needed for calculation of precision, recall, and $F_1$ are defined for binary input, to produce the count-preserving interpretation of the hierarchical measures, we need to use a count-preserving version of $TP$, $FP$, and $FN$. When comparing the number of predictions and true labels mapped to an ancestor:
\begin{equation}
TP_{c,d} = min(x_{c,d}, y_{c,d})
\label{eq:tp}
\end{equation}
\begin{equation}
FP_{c,d} = max(x_{c,d}-y_{c,d}, 0)
\label{eq:fp}
\end{equation}
\begin{equation}
FN_{c,d} = max(y_{c,d}-x_{c,d}, 0)
\label{eq:fn}
\end{equation}

\noindent Where $c$ represents a particular (ancestor) label, $d$ a specific document, $x_{c,d}$ and $y_{c,d}$ are the numbers of predicted and true prediction-level descendant labels of $c$ in $d$. The functions $min$ and $max$ return the minimum and maximum of two real numbers respectively. $TP$ (Equation \ref{eq:tp}) represents the overlap between the expected and predicted count of descendant codes of the ancestor $c$. $FP$ (Equation \ref{eq:fp}) and $FN$ (Equation \ref{eq:fn}) represent the over-prediction and under-prediction of these descendants, respectively.

\textbf{Remark:} Note that if $x_{c,d}$ and $y_{c,d}$ are binary, the results of these calculations are equivalent to those of standard $TP$, $FP$, $FN$.

 \begin{figure}
  \centering
  \includegraphics[width=\linewidth]{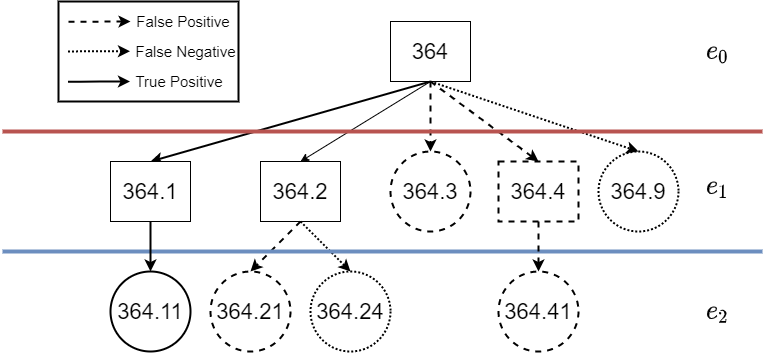}
  \caption{An example of hierarchical evaluation. Circular nodes represent leaf nodes (for non-hierarchical evaluation), borders of nodes represent set-based hierarchical evaluation, edges represent count-preserving hierarchical evaluation. Solid lines represent $TP$, dashed-lines represent $FP$, dotted lines represent $FN$. Levels of depth in the ontology ($e_0$, $e_1$, and $e_2$) are indicated with horizontal lines.}
  \label{fig:sub2}
\end{figure}

 \begin{table}[h]
 \small
\begin{tabular}{l||c|c|c||c|c|c}

 & $TP$ & $FP$ & $FN$ & $P$ & $R$  & $F_1$ \\
\hline
\hline
\textbf{Leaf-Only} & 1 & 3 & 2 & 25.0 & 33.3  & 28.4 \\
\hline
\hline
\textbf {Set-Based} & & & & &  & \\
$e_2$ & 1 & 2 & 1 & 33.3 & 50 & 40\\

$e_1$  & 2 & 2 & 1 & 50 & 66.7  & 57.1\\

$e_0$  & 1 & 0  & 0 & 100 & 100 & 100\\

\hline
Overall & 4 & 4 & 2 & 50 & 66.7  & 57.1\\
\hline
\hline
\textbf {CoPHE} & & & & &  & \\
$e_2$ & 1 & 2 & 1 & 33.3 & 50 & 40 \\

$e_1$  & 2 & 2 & 1 & 50 &  66.7  &  57.1\\

$e_0$  & 3 & 1 & 0 & 75 & 100  & 86\\

\hline
Overall & 6 & 5 & 2 & 54.5 & 75 & 63.1\\
\end{tabular}
\caption{Evaluation of the three representations presented in Figure \ref{fig:comparison}. In the case of Set-Based evaluation and CoPHE we present both the evaluation at each level of the ontology, and overall evaluation across levels. 
}
\label{tab:example}
\end{table}

 Suppose 4 descendants of code \emph{364} -- \emph{364.11}, \emph{364.21}, \emph{364.3}, and \emph{364.41} -- are predicted and 3 descendants of \emph{364} are true for document $\delta$ -- \emph{364.11}, \emph{364.24}, and \emph{364.9} (as seen in Figure \ref{fig:sub2}). In the count-preserving approach, according to our definitions, this results in $TP$ = 3, $FP$ = 1, $FN$ = 0 for the highest level (\emph{364}). The set-based version of hierarchical evaluation would only track if a prediction or gold standard path passes through a node, rather than their count (losing data on over- and under-prediction of each node's descendants). Here in the highest level (code \emph{364}), however, the four prediction paths and three gold standard paths are reduced to boolean \emph{True} for both prediction and gold standard. Hence the values for \emph{364} become $TP$ = 1, $FP$ = 0, $FN$ = 0. Note that due to this representation the over-prediction of descendants of \emph{364} is not captured (Figure \ref{fig:comparison} B). The binary and count-preserving evaluation can be produced for each individual code, aggregated over individual levels of depth of the label space or multiple consecutive layers of depth starting from the lowest. We refer to the latter aggregation in conjunction the with count-preserving approach as Count-Preserving Hierarchical Evaluation (CoPHE; Figure \ref{fig:comparison} C). The micro-averaged precision, recall and $F_1$ scores (both overall and per-level) of the example in Figure \ref{fig:comparison} are presented in Table \ref{tab:example}.

\section{Metric Analysis}
We produced evaluation results with CoPHE on three prior art models on MIMIC-III.\footnote{Since BERT based models are yet to be well adapted to MIMIC-III ICD coding \cite{gao2021,ji2021}, we leave BERT models for a future study.} CAML \cite{Mullenbach_Wiegreffe_Duke_Sun_Eisenstein_2018} is the first notable LMTC model developed on MIMIC-III whose contribution to the task was the introduction of label-wise attention. \citet{chalkidis2019large} proposed an alteration on CAML called BIGRU-LWAN, (here referred to as BGLWAN) swapping CAML's CNN encoder with a bidirectional GRU, and making use of zero-shot-motivated label description embeddings proposed by \citet{Rios_Kavuluru_2018}. \citet {Dong_Suarez-Paniagua_Whiteley_Wu_2020} further added label-wise word- and sentence-level attention and incorporated label embeddings pre-trained from the training labelsets (instead of the label descriptions) into the prediction and attention layers producing the HLAN model.

In the context of MIMIC-III, results tend to be presented on two label sets -- the 50 most frequent codes (\emph{top50}) and the full codeset (\emph{full}) -- this stems from the big-head long-tail distribution of labels in MIMIC-III with many labels being infrequent, or not appearing within the training set at all. We have used the pre-processing and  \emph{top50} codeset dataset split of \citet{Mullenbach_Wiegreffe_Duke_Sun_Eisenstein_2018}. Our results are averaged across 10 runs and serve as a comparative example for the evaluation metrics. 

We have applied CoPHE up to the lowest chapter levels (one above $e_0$), which we simply refer here to as the chapter level ($c$). This choice was made due to the level above $c$ already including the root of the ontology for some codes, leading to structural inconsistency.

\begin{table*}[t]
\centering
\begin{tabular}{l||c|c|c||c|c|c||c|c|c}
\multicolumn{1}{c}{}& \multicolumn{3}{c}{Standard Flat} & \multicolumn{3}{c}{Set-Based} & \multicolumn{3}{c}{CoPHE}\\
Model & $P$ & $R$ &  $F_1$ & $P$ & $R$ & $F_1$ & $P$ & $R$ &  $F_1$\\

\hline
CAML & 59.3 & 61.4 & 60.1  & 62.6 & 65.5 & 64.2 & 61.1 & 65.4  & 63.1 \\
BGLWAN & 68.4  & 57.6 & 62.5 & 71.8 & 61.2 & 66.0 & 70.5 & 61.4  & 65.1  \\
HLAN & 73.9 & 57.4 & 64.2 & 77.0 & 60.3 & 68.5 & 76.1 & 59.1 & 66.5 \\

\end{tabular}
\caption{ A comparison between flat evaluation and hierarchical evaluation -- Set-Based and CoPHE -- on three models from prior art using the \emph{top50} codeset. The results are micro-averaged across labels (and ancestors for hierarchical measures). Levels of hierarchy up to and including the lowest chapter level ($e_2$, $e_1$, $e_0$, $c$) are considered. Hierarchical measures report higher $F_1$ scores than the original flat measures. Furthermore, Set-Based evaluation $F_1$ scores are higher than those of CoPHE.
}
\label{tab:overall}
\end{table*}

We compared the flat metric results against the results of the set-based and CoPHE hierarchical measures on the three models (Table \ref{tab:overall}). The scores of CoPHE are higher than that of flat evaluation, showing that the measure is more lenient in its interpretation of errors. 

The scores of CoPHE are consistently lower than that of set-based evaluation, showing that CoPHE can capture more over- and/or under-prediction of the labels, as the True Positives of the ancestor codes are better manifested in CoPHE. If all (binary) True Positives within higher levels of the hierarchy reported by the set-based measure were True Positives in CoPHE (no over- or under-predictions), this would result in a greater proportion of True Positives for CoPHE than in the set-based measure, and hence higher scores (see Table \ref{tab:example} as an example). In contrast, the consistent lower CoPHE scores across the three methods imply the presence of over- and/or under-prediction within families of codes that is not captured by the set-based measure. 

The change of evaluation measure has not affected the ranking of the models on $F_1$. This is likely due to a lack of representation of the label space structure within the explored models. 

While hierarchical measures enrich the evaluation with structural information, they do not serve as a replacement of the flat measures, but rather should be viewed side-by-side. An approach may not surpass the state-of-the-art on the flat measures, whilst being better at modelling the structure of the label space based on the hierarchical score. Indication of better modelling of the structure should prompt further analysis of the model.

\section{Conclusion}
We have proposed the use of hierarchical evaluation measures in the LMTC task involving hierarchical label spaces and provided an example in the task of automated ICD coding. Unlike the approaches in prior art of ICD coding, which penalise all mispredictions equally, the proposed hierarchical evaluation measures adjust the penalty based on the performance on the ancestral levels. We have described a means to represent the hierarchy according to depth within the ontology. Finally we have proposed the use of count-preserving evaluation which captures data on both over and under-prediction in ancestral levels, as opposed to an existing set-based (binary) hierarchical evaluation approach.
\section{Future Work}
We intend to use the proposed hierarchical evaluation metrics alongside the flat metrics from prior art in our future experiments, particularly in the ones incorporating ontological structure within the model (similar to \citet{Rios_Kavuluru_2018}). The code for the proposed metric along with the level-based representation of the ontology have been made public in order to aid future work. The hierarchical evaluation can be applied to other LMTC tasks using different hierarchical label spaces. Finally, not all structured label spaces follow a tree structure -- it is important to explore the possibility of similar measures for ontologies with more generic graphs, \eg SNOMED CT.\footnote{\url{https://www.nlm.nih.gov/healthit/snomedct/index.html}}

\section*{Acknowledgements}
This work was supported by the United Kingdom Research and Innovation (grant EP/S02431X/1), UKRI Centre for Doctoral Training in Biomedical AI at the University of Edinburgh, School of Informatics, Health Data Research UK (HDR UK) National Text Analytics and Phenomics Projects, and the Wellcome Institutional Translation Partnership Award (PIII009).
The authors would like to thank Francisco Vargas and Andreas Grivas for their feedback on the visualisation of CoPHE. 
\bibliography{anthology,custom}
\bibliographystyle{acl_natbib}
\end{document}